\documentclass{article}

 \PassOptionsToPackage{numbers, compress}{natbib}
 \usepackage[final]{nips_2017}

\usepackage[utf8]{inputenc} 
\usepackage[T1]{fontenc}    
\usepackage[table]{xcolor}
\usepackage{hyperref}       
\usepackage{url}            
\usepackage{booktabs}       
\usepackage{amsfonts}       
\usepackage{nicefrac}       
\usepackage{microtype}      
\usepackage{amssymb}
\usepackage{pifont}

\usepackage{amsmath,amssymb,xspace,amsthm,Tabbing,bm}
\usepackage{mathtools}
\usepackage{amsthm}
\usepackage{listings}
\usepackage{multirow}
\usepackage{fullpage}
\usepackage{bbm}
\usepackage[export]{adjustbox}
\usepackage{wrapfig}
\usepackage{calrsfs}

\DeclareMathAlphabet{\pazocal}{OMS}{zplm}{m}{n}

\usepackage{times}
\usepackage{graphicx} 

\usepackage{natbib}

\usepackage{graphicx}
\usepackage{caption}
\usepackage{subcaption}

\usepackage{algorithm}
\usepackage{algorithmic}

\newtheorem{takeaway}{Take-Away}

\providecommand{\nor}[1]{\left\lVert {#1} \right\rVert}

\providecommand{\scal}[2]{\left\langle{#1},{#2}\right\rangle}

\providecommand{\scalT}[2]{\left\langle{#1},{#2}\right\rangle}

\newcommand{\R}{\mathbb R}
\providecommand{\scal}[2]{\left\langle{#1},{#2}\right\rangle}

\def\bit{\begin{itemize}}
\def\eit{\end{itemize}}

\def\ben{\begin{enumerate}}
\def\een{\end{enumerate}}

\def \P{\mathbb{P}}
\def \Q{\mathbb{Q}}
\def \F{\mathcal{F}}
\def \E{\mathbb{E}}
\def \L{\pazocal{L}} 
\def \X{\pazocal{X}} 
\def \om{\hat{\Omega}} 

\definecolor{myblue}{rgb}{0.82,0.95,1.0}

\title{Semi-Supervised Learning with IPM-based GANs: an Empirical Study}

\author{
Tom Sercu, Youssef Mroueh \\
   \texttt{tom.sercu1@ibm.com}, 
    \texttt{mroueh@us.ibm.com} \\
   AI Foundations, IBM Research  \\
   IBM T.J Watson Research Center \\
}

\begin{document}

\maketitle

\begin{abstract}
We present an empirical investigation of a recent class of Generative Adversarial Networks (GANs) using 
Integral Probability Metrics (IPM) and their performance for semi-supervised learning.
IPM-based GANs like Wasserstein GAN, Fisher GAN and Sobolev GAN have desirable properties in terms of theoretical understanding, training stability, and a meaningful loss.
In this work we investigate how the design of the critic (or discriminator) influences the performance in semi-supervised learning.
We distill three key take-aways which are important for good SSL performance:
(1) the $K+1$ formulation, (2) avoiding batch normalization in the critic
and (3) avoiding gradient penalty constraints on the classification layer.
\end{abstract}

\section{Introduction}
Most success in machine learning has been booked in the domain of supervised learning, when large quantities of labeled data are available.
However it is commonly agreed that this is not a sustainable way forward, as we want algorithms that can learn new tasks and generalize to new domains even when there is no or very little labeled data available.
One succesful approach to unsupervised and semi-supervised learning are Generative Adversarial Networks (GANs) \cite{goodfellow2014generative}, which are formulated as a min-max game between a generator $g_\theta(z)$ which defines an implicit density $\Q_\theta$ over $\X \subset \R^d$ from which we can sample, and a discriminator $D$ (or critic $f$) which measures a distance between the ``real'' $\P_r$ and ``fake'' $\Q_\theta$ distribution.
In order to use GANs for semi-supervised learning \cite{springenberg2015unsupervised,salimans2016improved} or unsupervised learning \cite{chen2016infogan}, typically the discriminator network is in some sense shared with a learned classifier $p(y|x)$, with $y \in \{1 \dots K\}$ the discrete label set. 
The underlying assumption is that features that are helpful towards telling real/fake apart, are also helpful towards classification.
Specifically the idea of the discriminator having ``$K+1$'' output directions, with the $K+1$\textsuperscript{th} ``fake'' direction competing with the K ``real'' directions \cite{salimans2016improved} has lead to strong empirical results \cite{dai2017good}.

Recently the distance metric between distributions became a topic of interest \cite{arjovsky2017towards,nowozin2016f,kaae2016amortised,mao2016least,arjovsky2017wasserstein,gulrajani2017improved,mroueh2017mcgan,mroueh2017fisher,li2017mmd,anon2017sobolev}.
We will focus on GANs with one class of distance metrics, Integral Probability Metrics or IPMs \cite{muller1997integral,sriperumbudur2009integral,sriperumbudur2012empirical}.
IPMs were first introduced in the GAN framework in Wasserstein GAN (WGAN) \cite{arjovsky2017wasserstein,gulrajani2017improved},
and also formed the basis for McGan \cite{mroueh2017mcgan}, Fisher GAN \cite{mroueh2017fisher} and Sobolev GAN \cite{anon2017sobolev}.
The IPM distance between $\P$ and $\Q$ is defined as:
\[ \sup_{f \in \F} \E_{x\sim \P} f(x)- \E_{x\sim \Q} f(x). \]
The IPM definition looks for a witness function (or ``critic'') $f(x)$, which maximally discriminates between samples coming from the two distributions.
This naturally fits the GAN idea: we can parametrize the critic with a neural network which takes the place of the discriminator in the GAN framework.
A crucial ingredient of the IPM metric is the function class $\F$, which defines how the critic $f(x)$ is bounded,
which in its turn \emph{defines the metric} being measured between distributions.

In WGAN \cite{arjovsky2017wasserstein,gulrajani2017improved} we approximate $\F$ the class of Lipschitz functions.
WGAN-GP \cite{gulrajani2017improved} uses a pointwise gradient norm penalty.
Fisher GAN \cite{mroueh2017fisher} introduces a tractable constraint on $\E_{x \sim \mu} f^2(x)$ which is enforced on samples from $\mu=\frac{\P+\Q}{2}$.
Sobolev GAN \cite{anon2017sobolev} introduces the tractable constraint on $\E_{x \sim \mu } \nor{\nabla_x f(x)}^2$ on the same $\mu$.
In this paper, we investigate which IPM formulations are amenable towards semi-supervised learning,
and whether we can leverage the $K+1$ formulation of classical JSD-based GANs \cite{salimans2016improved}.

\begin{figure}[t]
\centering
\includegraphics[width=0.75\linewidth]{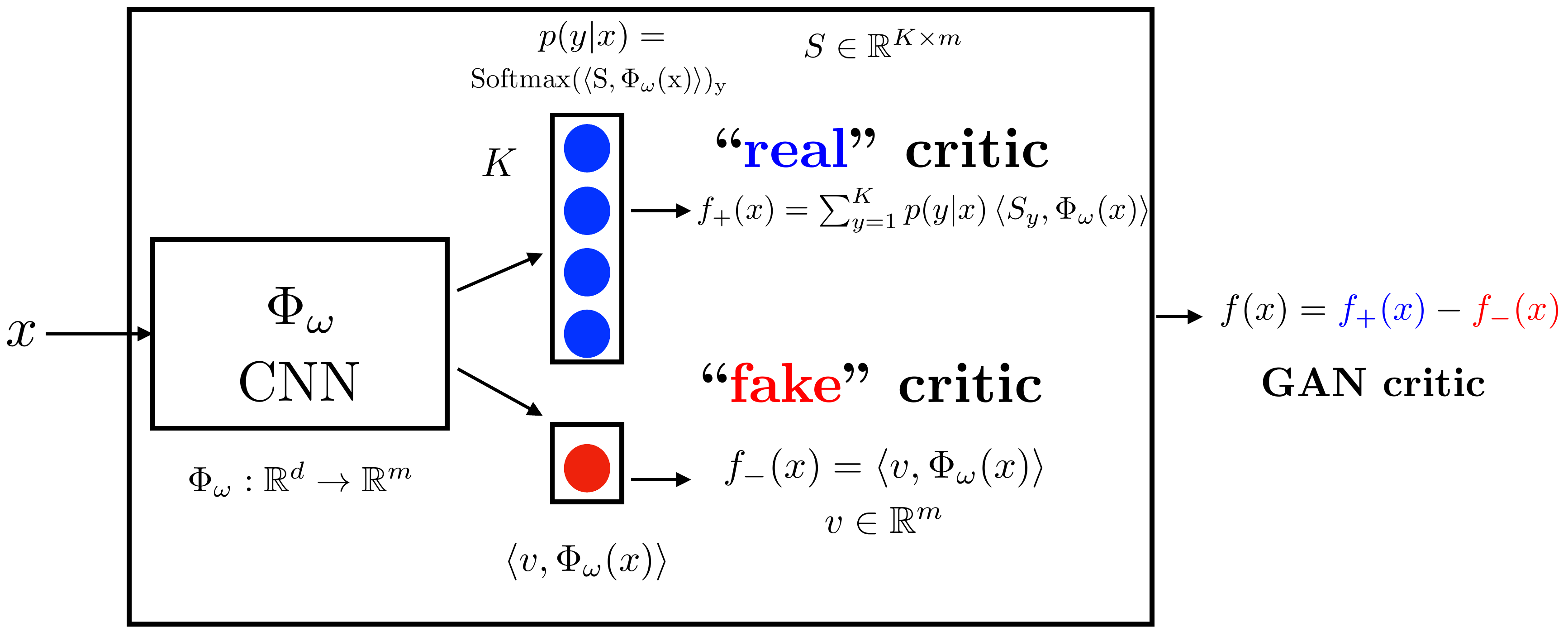}
\caption{``K+1'' parametrization of the IPM critic for semi-supervised learning.}
\label{fig:sslcritic}
\end{figure}

We will investigate the ``$K+1$ parametrization" of the critic as introduced in \cite{mroueh2017fisher} (See Figure \ref{fig:sslcritic}): 
\[ 
f(x) =\underbrace{ \sum_{y=1}^K p(y|x)\scalT{S_y}{\Phi_{\omega}(x)}}_{\bm{f_{+}}: \text{ ``real" critic }} - \underbrace{\scalT{v}{\Phi_{\omega}(x)}}_{\bm{f_-} :\text{``fake" critic}} 
\]
\vskip -1.0em
The training objective for critic and classifier now becomes:
\[
  \max_{S, \Phi_{\omega}, f } \L_{D} = \frac{1}{N} \left(\sum_{x \in \text{unl} } f(x) - \sum_{z \sim p_z} f(g_\theta(z)) \right) + \L^{\text{C}} (f_+, f_-, g_\theta) - \lambda_{CE} \sum_{(x,y) \in \text{lab} } CE (p(y|x),y )
\]
Where the constraint term $\L^\text{C}$ can contain the real and fake terms of the critic $f_+$ and $f_-$ separately.
Note that $p(y|x)=\rm{Softmax}(\scalT{S}{\Phi_{\omega}(x)})_y$ appears both in the critic formulation and in the Cross-Entropy term.
Intuitively this critic uses the $K$ class directions of the classifier $S_y$ to define the ``real'' direction, which competes with another $K+1$\textsuperscript{th} direction $v$ that indicates fake samples.

We empirically investigate the merit of this $K+1$ formulation in IPM-based GANs in Section \ref{sec:Kp1}.
In Section \ref{sec:constraint} we investigate on which parts of the critic $f, f_+, f_-$ to apply the constraint terms.
Finally, in Section \ref{sec:normalization}, we investigate the influence of activation normalization in the critic $f$, such as batch normalization (BN) \cite{ioffe2015batch} and layer normalization (LN) \cite{ba2016layer}.

\section{Experiments}
We will provide experimental results on CIFAR-10 \cite{cifar10} and SVHN \cite{netzer2011reading} using the CNN architectures as in \cite{mroueh2017fisher} for $g_\theta$ and $f$,
which is very close to the discriminator architecture used in \cite{salimans2016improved,dumoulin2016adversarially}.
Unless noted otherwise this CNN will have no normalization (LN or BN) in the critic, but $g_\theta$ always includes BN.
Similar to the standard procedure in other GAN papers we use 4k labeled samples for CIFAR-10, 1k labeled samples for SVHN, and do hyperparameter and model selection on the standard CIFAR-10 and SVHN validation split.
Hyperparameter details are in Appendix \ref{sec:hypers}.
We provide purely supervised baseline results of the critic architecture in Appendix \ref{sec:supervised}.


\subsection{$K+1$ formulations.} \label{sec:Kp1}
We show in Table \ref{tab:Kp1} the results for plain vs $K+1$ critic formulation.
``Plain $\scal{v}{\Phi_\omega(x)}$'' indicates the plain critic $f=\scal{v}{\Phi_\omega(x)}$ not interacting with the classifier.
For the $K+1$ formulations, the constraints act on the full critic, except for the combination where Fisher acts on $f$ but Sobolev is applied to $f_-$ (see next Section).

The third and sixth column of Table \ref{tab:Kp1} present results for a modified $K+1$ formulation $f^H = f_+^H - f_-$, where 
\[ 
f_+^{H}(x) 
= \sum_{y=1}^K p(y|x) \log( p(y|x) )
= \sum_{y=1}^K p(y|x)\scalT{S_y}{\Phi_{\omega}(x)} - \log Z_Y(x)
= f_+(x) - \log Z_Y(x)
\]

In this formulation, $f_+^H(x)$ is the negative entropy $-H[p(y|x)]$ of the classifier for a given $x$, which is what is being optimized as the sole objective in \cite{springenberg2015unsupervised}.
This $f_+^H$ is not sensitive to the magnitude of, or an additive bias to, the $\scalT{S_y}{\Phi_\omega(x)}$ because of the normalization constant $\log Z_Y(x) = \log( \sum_{y'=1}^K \exp \scalT{S_{y'}}{\Phi_\omega(x)})$.
This could either stabilize the training and shift more focus on the $K+1$\textsuperscript{th} direction $v$,
or could limit the effectiveness of $f_+$ in contrasting between real and fake samples - which makes it interesting to investigate.

We see that in the two succesful IPMs (Fisher and Fisher+Sobolev), the $K+1$ formulations give a strong gain
over the plain formulation.
Best results are obtained for the (unmodified) $K+1$ formulation and Sobolev + Fisher constraint.

\begin{takeaway}
The $K+1$ formulation is superior over the plain $f(x)=\scal{v}{\Phi_\omega(x)}$ formulation.
\end{takeaway}

\begin{table}[t]
\centering
\caption{Results for different critic formulations. Note that the $K+1$ formulation is better across the board (except for WGAN-GP because of the harmful gradient penalty on the full critic $f$, see Section \ref{sec:constraint}).
In gray we indicate all settings which failed to improve over straight-up supervised training with the small set of labeled samples (Appendix \ref{sec:supervised}).
}
\label{tab:Kp1}
\resizebox{\textwidth}{!}{ 
\begin{tabular}{| l | l |l | l | l | l | l |  }
  \toprule
                 & \multicolumn{3}{| c |}{CIFAR-10 (4k)}  & \multicolumn{3}{| c |}{SVHN (1k)} \\
                 &      Plain $\scal{v}{\Phi_\omega(x)}$ &             $K+1$ &  $K+1, f_+^H$ &     Plain $\scal{v}{\Phi_\omega(x)} $ &  $K+1$            &  $K+1, f_+^H$ \\
\midrule
WGAN clip  &  \cellcolor[gray]{0.9} $40.85$  & & & \cellcolor[gray]{0.9} $28.65$ &  & \\
WGAN-GP          &  \cellcolor[gray]{0.9} $44.64 \pm 0.61$ &  \cellcolor[gray]{0.9} $48.92 \pm 0.50$ &  \cellcolor[gray]{0.9} $48.81 \pm 0.41$ &                        $17.51 \pm 0.23$ &  \cellcolor[gray]{0.9} $31.59 \pm 1.75$ &  \cellcolor[gray]{0.9} $33.85 \pm 1.46$ \\
Fisher           &                        $18.95 \pm 0.32$ &                        $17.82 \pm 0.43$ &                        $17.12 \pm 0.11$ &                        $15.59 \pm 0.67$ &                         $9.46 \pm 0.32$ &                        $11.93 \pm 0.26$ \\
Sobolev          &  \cellcolor[gray]{0.9} $80.45 \pm 0.59$ &  \cellcolor[gray]{0.9} $45.35 \pm 0.62$ &  \cellcolor[gray]{0.9} $45.27 \pm 0.33$ &  \cellcolor[gray]{0.9} $80.46 \pm 0.62$ &  \cellcolor[gray]{0.9} $28.46 \pm 1.04$ &  \cellcolor[gray]{0.9} $26.00 \pm 0.43$ \\
Fisher + Sobolev &  $21.44 \pm 0.74$ &                        $\bm{16.29 \pm 0.42}$ &  $22.68 \pm 0.31$ &                        $13.23 \pm 0.53$ &                         $\bm{8.88 \pm 0.84}$ &                        $12.01 \pm 0.19$ \\
\bottomrule
\end{tabular}
} 
\vskip -0.13in
\end{table}

\subsection{$K+1$: constraints on $f$, $f_+$ or $f_-$} \label{sec:constraint}

\begin{table}[b]
\centering
\vskip -0.13in
\caption{ Four constraint combinations (GP, S, F, F+S), acting on either $f$, $f_-$, $f_+$.
  Failed settings are color-marked:
  blue indicates experiments that only include a penalty on $f_-$,
  gray marks the experiments where a gradient constraint was acting on the full critic $f$.
}
\label{tab:constraint}
\resizebox{0.40\textwidth}{!}{ 
\begin{tabular}{| l | r | r |}
\toprule
 & CIFAR-10 (4k) & SVHN (1k) \\
\midrule
\rowcolor[gray]{0.9}
$\om_{GP}(f)$           &          $49.28$ &       $34.05$ \\
\rowcolor{myblue}
$\om_{GP}(f_{-})$        &          $73.78$ &       $70.61$ \\
\rowcolor[gray]{0.9}
$\om_S(f)$            &          $45.55$ &       $26.15$ \\
\rowcolor{myblue}
$\om_S(f_{-})$         &          $71.38$ &       $20.04$ \\
$\om_F(f)$            &          $17.49$ &        $9.16$ \\
\rowcolor{myblue}
$\om_F(f_{-})$         &          $72.08$ &       $38.33$ \\
\rowcolor[gray]{0.9}
$\om_F(f) , \om_S(f)$       &          $45.55$ &       $29.30$ \\
$\om_F(f) , \om_S(f_-)$   &          $\bm{17.08}$ &        $8.27$ \\
\rowcolor[gray]{0.9}
$\om_F(f_-) , \om_S(f)$   &          $47.15$ &       $26.31$ \\
\rowcolor{myblue}
$\om_F(f_-) , \om_S(f_-)$ &          $68.62$ &       $37.50$ \\
\rowcolor[gray]{0.9}
$\om_F(f_+) , \om_S(f)$   &          $47.79$ &       $27.99$ \\
$\om_F(f_+) , \om_S(f_-)$ &          $17.12$ &        $\bm{8.01}$ \\
\bottomrule
\end{tabular}
} 
\end{table}

Similar to the notation in \cite{mroueh2017fisher,anon2017sobolev}
we write the Fisher and Sobolev constraint respectively as
$\om_{F}(f) = \frac{1}{N} \sum_{\tilde{x} \sim \mu} f^2(\tilde{x}) = 1$,
and
$\om_{S}(f) = \frac{1}{N} \sum_{\tilde{x} \sim \mu} \nor{\nabla_x f(\tilde{x})}^2 = 1$.
The WGAN-GP constraint is
$\nor{\nabla_x f(\tilde{x})}=1$ for points interpolated between real and fake, 
this is enforced through
$\om_{GP}(f) = \sum_{\tilde{x} \sim \mu_{GP}} (1 - \nor{\nabla_x f(\tilde{x})})^2$.
Both $\om_{F}$ and $\om_{S}$ are enforced with augmented lagrange multipliers with hyperparameters $\rho_F$ and $\rho_S$,
while $\om_{GP}$ is enforced with a large fixed penalty weight $\lambda_{GP}=10.0$.

Note now that the constraints can be enforced on either $f_+$, $f_-$, or the full critic $f = f_+ - f_-$.
To ensure that the critic is bounded, at least some constraint has to be acting on $f_-$ directly or through $f$,
while the boundedness of $f_+$ could in principle be ensured through the CE term on the small labeled set.
Note there will be non-trivial interaction between $\lambda_{GP}$, $\rho_F$, $\rho_{S}$ and $\lambda_{CE}$.

In Table \ref{tab:constraint} are results for the four different constraints and combinations acting on different parts of the critic.
We see that formulations with constraints only action on $f_-$ failed: clearly the CE term alone is not enough to constrain $f_+$.
Another important conclusion is that any combination where a gradient-norm constraint (Sobolev or WGAN-GP) is acting on the full critic $f$, the classifier is compromised:
in these settings it is impossible for the network to fit even the small labeled training set (heavy underfitting), causing bad SSL performance.

\begin{takeaway}
We need some form of constraint acting on both $f_-$ and $f_+$; the CE term alone is not enough to control $f_+$.
\end{takeaway}
\begin{takeaway}
Constraints including the gradient norm (WGAN-GP, Sobolev) should only act on $f_-$, otherwise the network underfits.
\end{takeaway}

\subsection{How to normalize the critic} \label{sec:normalization}
\begin{table}[t]
\centering
\caption{How to normalize the critic. We see BN performs signficantly worse than other options for Fisher GAN (and is incompatible with GP/Sobolev).
The layernorm formulation with singleton $\mu,\sigma^2$ statistics are superior to statistics per feature map.
Fisher GAN benefits from layernorm, while in Sobolev+Fisher no normalization is prefered - this gives the overall best result.
}
\label{tab:normalization}
\resizebox{\textwidth}{!}{ 
\begin{tabular}{lllll}
\toprule
     &  CIFAR-10 (4k) &                   &     SVHN (1k) &                  \\
IPM Definition &            Fisher &  Fisher + Sobolev &           Fisher & Fisher + Sobolev \\
$f$ normalization         &                   &                   &                  &                  \\
\midrule
Batch Normalization       &  $24.14 \pm 0.26$ &                   & $10.76 \pm 0.28$ &                  \\
LN ($\mu,\sigma^2 \in \R^{1\times 1\times 1}$) ($\bm{g}, \bm{b} \in \R^{C\times 1 \times 1}$) 
&  $\bm{16.45 \pm 0.42}$ &  $16.79 \pm 0.11$ &  $8.78 \pm 0.68$ &  $9.02 \pm 0.29$ \\
LN ($\mu,\sigma^2 \in \R^{1\times 1\times 1}$) ($\bm{g}, \bm{b} \in \R^{1\times H \times W}$) 
&  $16.81 \pm 0.36$ &  $17.24 \pm 0.34$ &  $\bm{8.55 \pm 0.29}$ &  $8.70 \pm 0.67$ \\
LN ($\mu,\sigma^2 \in \R^{C\times 1\times 1}$) ($\bm{g}, \bm{b} \in \R^{C\times 1 \times 1}$) 
                &  $20.44 \pm 0.24$ &  $20.09 \pm 0.43$ &  $9.18 \pm 0.12$ &  $9.19 \pm 0.21$ \\
LN ($\mu,\sigma^2 \in \R^{C\times 1\times 1}$) ($\bm{g}, \bm{b} \in \R^{1\times H \times W}$) 
                &  $19.93 \pm 0.34$ &  $19.85 \pm 0.29$ &  $9.69 \pm 0.16$ &  $9.29 \pm 0.16$ \\
                No Normalization &  $17.73 \pm 0.56$ &  $\bm{16.33 \pm 0.15}$ &  $9.06 \pm 0.46$ &  $\bm{8.31 \pm 0.60}$ \\
\bottomrule
\end{tabular}
} 
\end{table}

In the original DCGAN \cite{radford2015unsupervised} architecture, batch normalization (BN) \cite{ioffe2015batch} was a crucial ingredient,
both in the generator and discriminator.
Even though the original WGAN still relies on BN,
both WGAN-GP, Fisher GAN, and Sobolev GAN report strong results without any layerwise normalization.
When the constraint involves a norm of the gradient, BN is problematic since it couples the different samples in the batch.
Here we investigate as alternative to BN either layer normalization (LN) \cite{ba2016layer} or having no normalization in the critic.

One important detail which usually glossed over, is how layernorm exactly is extended to the convolutional setting,
where the activations at a given layer are $\in \R^{C \times H \times W}$.
Specifically we need to decide whether we will accumulate the statistics mean $\mu$ and variance $\sigma^2$ either into a singleton ($\in \R^{1\times 1\times 1}$), or separate per feature map ($\in \R^{C\times 1\times 1}$).
Similarly, we need to decide whether we will parametrize the scale $\bm{g}$ and bias $\bm{b}$ separate per feature map ($\in \R^{C\times 1\times 1}$), or separate per pixel ($\in \R^{1\times H\times W}$).
In the above, we used broadcasting notation, meaning that the singleton dimensions will be expanded to perform the elementwise operations.
For reference, in batch normalization for convolutional networks both mean $\mu$, variance $\sigma^2$, scale $\bm{g}/\gamma$ and bias $\bm{b}/\beta$ are collected separately per feature map, i.e. $\in \R^{C\times 1\times 1}$.
Implementation-wise we follow \cite{ren2016normalizing} in adding a small $\epsilon$ \emph{inside} the square root in $\sqrt{\sigma^2+\epsilon}$.
The results in Table \ref{tab:normalization} lead us to conclude:

\begin{takeaway}
Avoid batchnorm, definitely when constraining the gradient norm in objective, but it also hurts for Fisher GAN!
\end{takeaway}
\begin{takeaway}
The layer normalization formulation with singleton stats ($\mu,\sigma^2 \in \R^{1\times 1\times 1}$) and parameters per feature map ($\bm{g}, \bm{b} \in \R^{C\times 1 \times 1}$) is superior.
Fisher GAN benefits from this LN, while for Sobolev+Fisher no normalization is better.
\end{takeaway}

\section{Conclusion}
We empirically investigated how different types of IPM-based Generative Adversarial Networks 
can be used for semi-supervised learning.
A comparison with literature results is given in Appendix~\ref{sec:literature}.
Our main conclusions are (1) the $K+1$ formulation works, (2) batch normalization should be avoided, also in Fisher GAN,
and (3) gradient penalty constraints should act on $f_-$ only, not on the full critic which includes the classifier $p(y|x)$.

\bibliographystyle{unsrt}
\bibliography{refs}

\newpage
\appendix
\section{CIFAR-10: Comparison against literature}
\label{sec:literature}

\begin{table}[ht!]
\vskip -0.13in
\centering
\caption{
CIFAR-10 error rates for varying number of labeled samples in the training set.
Mean and standard deviation computed over 5 runs.
We only use the $K+1$ formulation of the critic.
Note that we achieve strong SSL performance without any additional tricks,
and even though the critic does not have any batch, layer or weight normalization.
\label{tab:literature} }
\resizebox{\textwidth}{!}{
\begin{tabular}{@{}lllll@{}} \toprule
Number of labeled examples & 1000 & 2000 & 4000 & 8000 \\
Model & \multicolumn{4}{c}{Misclassification rate} \\ \midrule
CatGAN \citep{springenberg2015unsupervised}    &              &      & $19.58$ &   \\
FM \citep{salimans2016improved} & $21.83 \pm 2.01$ & $19.61 \pm 2.09$ & $18.63 \pm 2.32$ & $17.72 \pm 1.82$ \\
ALI \citep{dumoulin2016adversarially}          & $19.98 \pm 0.3$ & $19.09 \pm 0.15$ & $17.99 \pm 0.54$ & $17.05 \pm 0.50$ \\
Tangents Reg \citep{kumar2017improved} & $20.06 \pm 0.5$    &        & $16.78 \pm 0.6$  &  \\
$\Pi$-model \citep{laine2016temporal} * &                   &        & $16.55 \pm 0.29$  &  \\
VAT  \citep{miyato2017virtual}        &                     &        & $14.87 $          &  \\
Bad Gan \citep{dai2017good} *         &                     &        & $14.41 \pm 0.30$ &  \\
\midrule
\midrule
Fisher, batch norm  \citep{mroueh2017fisher} & $36.80$ &  $30.43$ &  $24.35$ &  $21.35$ \\
Fisher, layer norm \citep{mroueh2017fisher} & $19.74 \pm 0.21$ &  $17.87 \pm 0.38$ &  $16.13 \pm 0.53$ &  $14.81 \pm 0.16$ \\
Fisher, no norm \citep{mroueh2017fisher} & $21.15 \pm 0.54$ &  $18.21 \pm 0.30$ &  $16.74 \pm 0.19$ &  $14.80 \pm 0.15$ \\ 
Fisher+Sobolev, no norm \citep{anon2017sobolev}  & $20.14 \pm 0.21$ &  $17.38 \pm 0.10$ &  $15.77 \pm 0.19$ &  $14.20 \pm 0.08$ \\
\bottomrule
\end{tabular} }
\end{table}

\section{Hyperparameters}
\label{sec:hypers}

Unless noted otherwise, we use Adam with learning rate $\eta=2\mathrm{e}{-4}$, $\beta_1=0.5$ and $\beta_2=0.999$, both for critic $f$ (with and without BN / LN) and Generator (always with BN).
We use $\lambda_{CE}=1.5$ for the $K+1$ formulations (when the CE objective competes with the IPM objective),
but found a lower $\lambda_{CE}=0.1$ optimal for the regular ``plain'' formulation where the classification layer $S$ doesn't interact with the classifier.
We train all CIFAR-10 models for 350 epochs and SVHN models for 100 epochs.
We used some L2 weight decay: $1\mathrm{e}{-6}$ on $\omega, S$ (i.e. all layers except last) and $1\mathrm{e}{-3}$ weight decay on the last layer $v$.
We have $\rho_F=1\mathrm{e}{-7}$, and $\rho_S=1\mathrm{e}{-8}$. 
We keep critic iters $n_c=2$, but we noticed marginally better results with $n_c=1$ for the last experiment in Table \ref{tab:literature}.
For WGAN-GP we use the WGAN-GP defaults of $\lambda_{GP}=10.0$, $n_c=5$, learning rate $\eta_D=\eta_G=1\mathrm{e}{-4}$.

\section{Purely supervised baseline}
\label{sec:supervised}
Misclassification rate of CIFAR-10 and SVHN, for either the typical small labeled set,
or using all labels.
This provides a baseline result for the performance of our critic CNN $f$.
Here, results with BN are better than with LN, which is slightly better than without normalization.

\begin{tabular}{lllll}
\toprule
{}  &           CIFAR-10 &                   &              SVHN &                  \\
\# labeled &              4k &               all &              1k &              all \\
\midrule
BN           &  $34.17 \pm 0.16$ &  $11.47 \pm 0.10$ &  $21.67 \pm 0.39$ &  $4.11 \pm 0.04$ \\
LN ($\mu,\sigma^2 \in \R^{1\times 1\times 1}$) ($\bm{g}, \bm{b} \in \R^{C\times 1 \times 1}$) 
  &  $38.85 \pm 0.37$ &  $14.01 \pm 0.25$ &  $21.57 \pm 0.36$ &  $3.81 \pm 0.10$ \\
No Normalization                &  $39.78 \pm 0.28$ &  $14.25 \pm 0.25$ &  $21.89 \pm 0.36$ &  $3.92 \pm 0.08$ \\
\end{tabular}

\end{document}